\documentclass[sigconf, nonacm, anonymous=false]{acmart}

\usepackage{graphicx}
\usepackage{tikz}
\usepackage{pgfplots}
\usepackage{fancyvrb}
\usepackage[symbol]{footmisc}
\usepackage{xcolor}
\usepackage{lineno}

\AtBeginDocument{%
  \providecommand\BibTeX{{%
    \normalfont B\kern-0.5em{\scshape i\kern-0.25em b}\kern-0.8em\TeX}}}

\setcopyright{acmcopyright}
\copyrightyear{2018}
\acmYear{2018}
\acmDOI{XXXXXXX.XXXXXXX}

\acmConference[Conference acronym 'XX]{Make sure to enter the correct
  conference title from your rights confirmation emai}{June 03--05,
  2018}{Woodstock, NY}
\acmBooktitle{Woodstock '18: ACM Symposium on Neural Gaze Detection,
 June 03--05, 2018, Woodstock, NY} 
\acmPrice{15.00}
\acmISBN{978-1-4503-XXXX-X/18/06}

\begin{document}

\title{\textit{The first step is the hardest}: Pitfalls of Representing and Tokenizing Temporal Data for Large Language Models}

\author{Dimitris Spathis }\authornote{Corresponding author. Email: \href{mailto:dimitrios.spathis@nokia.com}{dimitrios.spathis@nokia.com}} 
%\email{dimitrios.spathis@nokia.com}
\affiliation{%
  \institution{Nokia Bell Labs}
  \streetaddress{P.O. Box 1212}
  \city{Cambridge}
  \country{UK}
}

\author{Fahim Kawsar}
%\email{dimitrios.spathis@nokia.com}
\affiliation{%
  \institution{Nokia Bell Labs}
  \streetaddress{P.O. Box 1212}
  \city{Cambridge}
  \country{UK}
}
\vspace{0.9cm}

\renewcommand{\shortauthors}{Spathis}

\begin{abstract}
Large Language Models (LLMs) have demonstrated remarkable generalization across diverse tasks, leading individuals to increasingly use them as personal assistants and universal computing engines. Nevertheless, a notable obstacle emerges when feeding numerical/temporal data into these models, such as data sourced from wearables or electronic health records. LLMs employ tokenizers in their input that break down text into smaller units. However, tokenizers are \textit{not} designed to represent numerical values and might struggle to understand repetitive patterns and context, treating consecutive values as separate tokens and disregarding their temporal relationships. Here, we discuss recent works that employ LLMs for human-centric tasks such as in mobile health sensing and present a case study showing that popular LLMs tokenize temporal data incorrectly. To address that, we highlight potential solutions such as prompt tuning with lightweight embedding layers as well as multimodal adapters, that can help bridge this "modality gap". While the capability of language models to generalize to other modalities with minimal or no finetuning is exciting, this paper underscores the fact that their outputs cannot be meaningful if they stumble over input nuances.

\end{abstract}




\maketitle

\section{Introduction}

In recent years, Large Language Models (LLMs) --also known as foundation models \cite{bommasani2021opportunities}-- have garnered attention for their ability to generalize across a wide array of tasks. From natural language understanding to generating creative text, these models have demonstrated their prowess, often mimicking human-like performance. Unlike previous AI models, prompting enables example-based learning without any additional training (a concept known as in-context learning \cite{openai2023gpt}). As their capabilities have grown, so too the idea of employing LLMs as general-purpose assistants has emerged, promising enhanced user experiences through personal devices.

However, amidst these advancements, a significant challenge arises when attempting to use existing LLMs with numerical and temporal data, particularly those originating from mobile sensors. These data sources, often representing fine-grained behavioral and physiological states, introduce a complex layer of information that cannot be easily "translated" into text. LLMs, designed primarily for processing natural language, exhibit remarkable efficiency in representing textual input through specialized tokenizers, effectively dividing text into manageable units. Yet, as we start to integrate LLMs with temporal data, the complexities of this union become evident.

The tokenizers employed by LLMs appear to stumble when grappling with numerical inputs. Repetitive patterns, an inherent characteristic of time-series data, can confound tokenizers, leading to the unintentional fragmentation of continuous sequences into disjointed tokens. Consequently, the temporal relationships that underpin such data may be lost in translation, potentially undermining the very essence of the information being processed.

In this context, this paper delves into the nuances and obstacles that emerge when LLMs are confronted with the task of representing and tokenizing temporal data. We focus on the interplay between numerical and textual information, uncovering the potential pitfalls that can hamper the effective utilization of LLMs in scenarios where temporal context is important. Last, we discuss potential solutions from the rapidly growing area of parameter-efficient transfer learning and multimodal adapters that could enable better integration of non-textual data into LLMs.

\section{Tokenization in Language Models}

Tokenization is a fundamental process underpinning the operation of LLMs. It involves the division of input and output texts into smaller, manageable units known as tokens. These tokens serve as the building blocks of language comprehension, enabling LLMs to process and generate text effectively. Tokens can encompass a variety of segments, such as characters, words, subwords, or symbols, depending on the chosen tokenization scheme. 

The purpose of tokenization extends beyond mere text segmentation. It enables LLMs to handle diverse languages, vocabularies, and formats while mitigating computational and memory demands. The quality and diversity of generated text are influenced by tokenization, as it shapes the meaning and context of individual tokens. Tokenization strategies vary and can be rule-based, statistical, or learnable in nature, adapting to the complexity and variability of the input texts.

One of the prominent subword tokenization methods used in Transformers, the architecture underlying many LLMs, is Byte-Pair Encoding (BPE) \cite{sennrich2016neural}. OpenAI, for instance, employs BPE in its GPT-based models, which include around 50,000 tokens. BPE operates by iteratively merging the most frequently occurring pairs of characters or bytes into a single token until a predefined number of tokens or vocabulary size is reached. This approach is especially useful for accommodating rare or previously unseen words, resulting in more compact and consistent representations of text.

In addition to BPE, other tokenization techniques include WordPiece \cite{schuster2012japanese} and SentencePiece \cite{kudo2018sentencepiece}. WordPiece tokenization segments text into words or subwords, often leading to smaller vocabularies and efficient processing. SentencePiece, on the other hand, extends tokenization to the sentence level, empowering models to handle morphologically rich languages with complex sentence structures, such as Chinese and Japanese.

As an example, let's consider the word \texttt{"playing"} to illustrate how BPE works. First, the algorithm identifies the most frequent pair of characters, that according to our hypothetical corpus is \texttt{"ay"}. It then merges \texttt{"ay"} to create a new token: \texttt{"pl"+"ay"= "play"}. Our new updated vocabulary now is \texttt{"play"} and \texttt{"ing"}. In this example, by breaking down words into subword units and merging frequent pairs, BPE captures meaningful components of words, even those that might not have appeared in the original vocabulary. 

\section{Where do tokenizers struggle?
}

Despite its importance, tokenization introduces various challenges and open problems that impact the quality, interpretation, and usability of LLMs' outputs. While tokenization schemes like Byte-Pair Encoding (BPE) have been successful in segmenting text into meaningful units, several issues persist, highlighting areas where further research is needed.

\textbf{Case sensitivity}. Tokens are assigned numerical identifiers within a tokenizer's vocabulary. However, cases of words are treated as separate tokens, leading to inconsistencies in representation. For instance, \texttt{"good"} and \texttt{"Good"} are encoded as distinct tokens, hampering the LLM's ability to understand text, particularly in scenarios where capitalization matters.

\textbf{Trailing whitespace}. Tokens with whitespace are treated as separate entities; for instance, tokens such as \texttt{"the first step is "} and \texttt{"the first step is"} exhibit different representations, impacting the probabilities of subsequent tokens. This behavior introduces subtle biases that affect the model's language generation process and response characteristics.

\textbf{Digit chunking}. The tokenization of numerical values poses challenges, particularly when digits are inconsistently chunked. Numbers like \texttt{"480"} might be tokenized as a single unit, while \texttt{"481"} and  \texttt{"482"} are split into two tokens. This inconsistency not only hampers the model's mathematical capabilities but also introduces complexities when dealing with consecutive values and temporal dependencies. 

\begin{Verbatim}[frame= single, commandchars=\\\{\}]
Input         → Token IDs
480\textcolor{lightgray}{,} 481\textcolor{lightgray}{,} 482 → 22148, \textcolor{lightgray}{11}, 4764, 16, \textcolor{lightgray}{11}, 4764, 17
\end{Verbatim} 

\textbf{Integers}. The particular case of integer tokenization was investigated in a recent work \cite{berenIntegerTokenization}. In a series of simulations with popular LLMs, it was found that the tokenization of integers lacks a coherent decimal representation, leading to a fragmented approach where even basic mathematical operations require memorization rather than algorithmic processing. On the other hand, non-unique integers are tokenized through arbitrary and inconsistent chunking, complicating multi-digit arithmetic operations. The irregular division of numbers into tokens varies even for adjacent integers, demanding models to memorize arbitrary tokenization patterns. Notably, tokenization is influenced by the temporal distribution of data. Common dates, mostly in the 20th century, are assigned unique tokens, revealing the problematic interplay between tokenization and the characteristics of the training dataset.

\textbf{Floating-point Numbers}. These numbers, often representing decimal values, can be challenging to tokenize consistently. The representation of floating-point numbers can vary based on factors such as precision, scientific notation, and rounding errors. Tokenization of floating-point numbers involves decisions on whether to treat them as a single token or break them down into constituent parts. For instance, consider the number \texttt{"3.14159"}, which might be tokenized as four different chunks by popular LLMs: \texttt{"3" + "." + "14" + "159"}.

\begin{Verbatim}[frame= single, commandchars=\\\{\}]
Input   → Token IDs
3\textcolor{lightgray}{.}14159 → 18, \textcolor{lightgray}{13}, 1415, 19707
\end{Verbatim} 

\textbf{Arithmetic reasoning}. LLMs exhibiting inconsistent tokenization of numbers struggle with addition tasks that involve two-digit numbers and completely falter in adding larger numbers \cite{nogueira2021investigating}. Specifically, their accuracy drops to zero for digits totalling five or more. This deficiency is attributed to the absence of a systematic approach to tokenizing individual digits. For instance, \texttt{"1234"} might be segmented into \texttt{"12"}, \texttt{"34"}, while \texttt{"5678"} could be split into \texttt{"5"} and \texttt{"678"}. Consequently, the model must grasp that a token's embedding could signify either a single digit or multiple digits, introducing complexity when mapping embeddings to varying-digit numbers. This irregular mapping challenge hinders the model's ability to accurately establish associations between embeddings and numbers of varying digit lengths. 

\textbf{Model-specific behavior}. Tokenization is not universally consistent across different language models. Even when employing the same underlying method, different models can have distinct token representations. This model-specific nature of tokenization poses challenges for interoperability, pre-processing, and multi-modal modeling. It necessitates careful consideration when assessing various LLMs for real-world applications.

\section{a case study on tokenizing sensing data}

\begin{figure*}[h]
  \centering
  \includegraphics[width=0.79\linewidth]{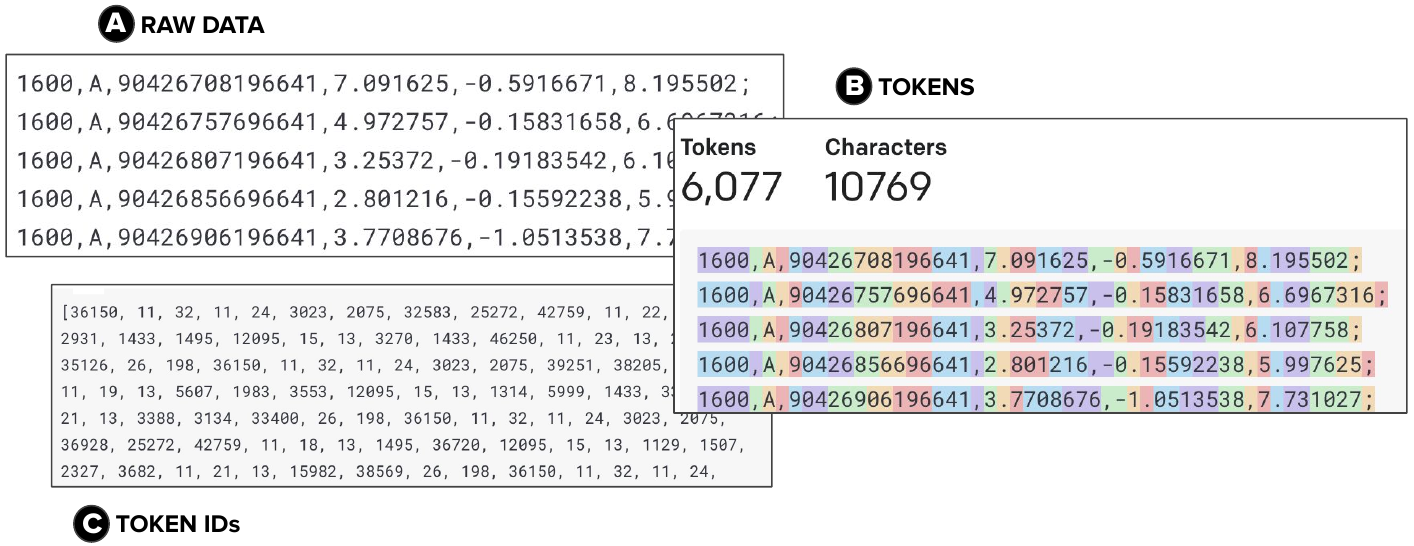}
  \caption{\textbf{Example of incorrect tokenization of a mobile sensing dataset}. \textmd{The first 200 rows (a) of WISDM activity recognition dataset (only the first 5 are visible) as it is processed (b) into tokens (c). The timestamps, acceleration values, and negative signs are split into subtokens that are not meaningful to models. Each row includes the Participant ID, activity label, UNIX timestamp, and  x, y, z acceleration.}}
 \label{fig:wisdm}
\end{figure*}

To showcase some of these challenges in the real world, we encode the first 200 rows of a popular mobile sensing dataset (WISDM \cite{weiss2019smartphone}) via the OpenAI Tokenizer that powers ChatGPT \footnote{The public OpenAI tokenizer was used in all examples in sections 3 and 4: \url{https://platform.openai.com/tokenizer}}. Each row contains the Participant ID, the activity label code, a UNIX timestamp, and the 3 accelerometer axes (x, y, z). As is common, models learn the temporal dependencies across the 3 axes in order to predict whether users perform activities such as running or walking. For illustration purposes, we choose not to filter or pre-process any information of the dataset.

In Figure \ref{fig:wisdm}, we show the raw input data along with the tokenized output and the token IDs. Affirming the concerns outlined in the previous section, we note the model's tendency to divide timestamps into multiple tokens, potentially rendering temporal dependencies across different rows irrelevant. With regard to the actual sensor data, each accelerometer value, which is naturally represented as a floating point number, is also split into multiple tokens. Adding to the complexity, the negative sign in front of the sensor values is grouped with the comma separator, thereby disregarding the direction of the sensor reading.

We argue that this representation cannot allow for any meaningful inference from LLMs. As we discuss in the following section, all existing preliminary works that attempt such inferences, have resorted to extensive filtering, such as downsampling, aggregating the signal, as well as rounding. As a result, raw data is rarely provided without context to models, with a lot of effort put into constructing well-crafted textual templates (prompts), in order to overcome this modality mismatch.

\section{large language models in sensing}

Researchers have started experimenting with treating LLMs as universal pre-trained models \cite{lu2021pretrained}, despite any modality gap. In this section, we will discuss preliminary works that feed temporal data into LLMs in order to improve applications in ubiquitous computing. We also note that while there are plenty of works that borrow the Transformer/GPT architecture and re-train from scratch \cite{vaid2023foundational, gaudilliere2021generative}, \textit{here we focus on the paradigm of directly feeding temporal data to pre-trained LLMs.}

\textbf{Self-supervised learning}. Despite the popularity of LLMs in language tasks, the established training paradigm of neural networks for ubiquitous computing tasks still involves a combination of convolutional, recurrent, and attention layers \cite{ma2019attnsense}. For example, in sensor-based Human Activity Recognition (HAR), a common workflow involves a sliding window which splits the input signal into fixed-sized sequences of 3 channels (accelerometer axes) after some normalization (usually standard scaling). Due to inherent difficulties in collecting large labeled datasets, self-supervised learning has proved to be particularly effective in mobile and health-sensing tasks \cite{saeed2019multi, haresamudram2022assessing, tang2021selfhar, spathis2022breaking}.

\textbf{Lack of pre-trained models}. Self-supervised learning learns robust representations through proxy tasks that operate on data without annotations. However, even large unlabeled data of that kind is hard to encounter because it is not publicly available on the web, unlike images or text. Another crucial limiting factor is the number of potential modalities; for example, aligned and paired data of movement, heart-rate, audio, and sleep patterns might be impossible to collect and share at scale due to privacy concerns. As a result, available pre-trained models in areas such as physical activity or heart health are limited in size and generalization capabilities, compared to popular foundation models \cite{yuan2022self, spathis2021self}.

\textbf{From language to physiological data}. One comprehensive albeit preliminary work in this area suggested that LLMs are few-shot health learners for a range of consumer health tasks \cite{liu2023large}. Specifically, the authors show that an LLM can be tuned to achieve strong performance on tasks like heart rhythm classification, activity recognition, calorie estimation, and predicting stress levels and mental health scores from Fitbit data. The paper also explores the effects of providing textual context versus solely numerical data as input to the LLM. Across almost all tasks, including more contextual information like \textit{"Classify the given Interbeat Interval sequence in ms"} led to improved performance over just feeding in the raw numbers. This suggests that contextual hints help anchor the numerical data and allow the LLM to leverage its domain knowledge more effectively. 

%Regarding the base model, the authors employ the proprietary 24B PaLM model which uses the SentencePad tokenizer with a 256K-sized token vocabulary \cite{chowdhery2022palm}.

Considering the above experimental setup, three evaluation protocols were followed:
\begin{itemize}
 \item \textbf{Zero-shot inference}: The LLM makes predictions directly on the test data without any additional tuning or training. This tests the out-of-the-box capabilities of the model but performance was poor on most tasks.
\item \textbf{Prompt engineering}: A few sample prompts are provided to the LLM before testing to demonstrate the task structure. This guides the model but doesn't update any parameters. Performance improved over zero-shot but was still limited.
\item \textbf{Prompt tuning}: A trainable prompt embedding is added to the input of the LLM and trained on 3-25 examples per task. This tuning dramatically improved performance across tasks compared to the other methods.
\end{itemize}

%Prompt tuning provides the best performance by requiring additional training that forces the model to adapt to the new tasks. It also requires some labeled examples while zero-shot does not. Zero-shot has the advantage of no additional data requirements but lacks the ability to adapt or improve. Prompt engineering strikes a balance by guiding the model but not training, though performance gains are limited. Overall, the experiments highlight tradeoffs between generalization. 
While acknowledging the promise of these results, there is limited information on any pre-processing or tokenization. We assume that once in textual prompt format, the time-series data is directly fed into the base model \cite{chowdhery2022palm} without any further vectorization or embedding. We hypothesize that zero-shot and prompting underperform because of the tokenization and representation challenges we discuss in sections 3 and 4. In section 6, we expand on the importance of prompt tuning and compare it to other approaches that could alleviate modality mismatches between numbers and language.

\textbf{From language to gestures}. Other preliminary works followed similar experimental setups. In a zero-shot scenario, the popular GPT4 model was fed in hand gesture data through light and vibration sensors \cite{sooriya2023poster}. The authors report high accuracies using prompt engineering templates such as: \textit{"I use an RGB Light and Proximity Sensor to detect the gestures Single air tap, Double air Tap, and Triple air Tap. [..] Given below are sample data collected at three instances for each gesture: [Numbers]"}. Despite the promising result, this work employed a particularly small dataset and considered zero-shot prompting only. It also did not report baseline model accuracy, making it hard to gauge the difficulty of the task.

\textbf{From language to forecasting}. PromptCast also followed a similar scenario by framing timeseries forecasting as a language generation problem and introduced a new prompt template dataset covering weather, energy, and human mobility forecasting \cite{xue2022prompt}. Experiments compared state-of-the-art numerical forecasting methods like Informer, Autoformer, and FEDformer against language models like T5, BART, and BigBird. The results showed that language models are competitive to numerical methods when fine-tuned on the datasets, with Bigbird, BART, and RoBERTa being the top performers. The language models also showed stronger generalization under zero-shot transfer. We attribute the success of such a setup to careful dataset creation and the short sequence length (15 timesteps) used in the prompts. We expect that by incorporating prompt-tuning or other adapter layers, these models would further improve over purely numerical models.

\section{How to better integrate temporal data into LLMs?
}

Despite the encouraging results, we see that these preliminary works have to "ground" the LLMs in numerical data using verbose hand-engineered prompts and extensive aggregation of timeseries. In other words, they rely predominantly on the power of text to contextualize numerical data. This paper argues that we will only unlock the real value of temporal data of that kind, once we map it to a more meaningful representation. While modality mappings between images and text are well-studied \cite{liang2022mind, shi2023towards}, the link of timeseries to text remains virtually unexplored. 

\textbf{Prompt tuning}. Parameter-Efficient Fine-Tuning (PEFT) has emerged as a potential solution to this problem. Traditional neural network fine-tuning would involve a pre-trained network that is fixed (or frozen) and new trainable layers that are added to the end of the network in order to tailor it to a specific task. Instead, PEFT involves prompting or lightweight layers that are added usually to the input (prompt-tuning) or selected layers throughout the model (prefix-tuning).

In general, prompt tuning refers to techniques that change the LLM prompt to achieve better results. Directly changing the input content is known as \textit{hard} prompt. More interestingly, \textit{soft} prompting involves prepending a trainable tensor to the input which can then be optimized to a downstream task via backpropagation \cite{lester2021power}. As seen in the previous section, this approach is particularly effective because the prompt layer encodes information that assists the model in comprehending timeseries data that is absent in the original frozen LLM \cite{liu2023large}. This is significantly more parameter-efficient than full fine-tuning which becomes prohibitive as the model size and the number of tasks grow.

On a similar note, modules known as \textit{Adapters} insert small trainable blocks to each layer of the pre-trained network with only these blocks being fine-tuned. For example, a popular method, LoRa, injects trainable low-rank matrices into transformer layers to approximate the weight updates \cite{hu2021lora}. While these approaches have similar goals, here we are interested in prompt-tuning because it operates on the input space and can effectively steer the model to process timeseries data \footnote{A unified view across all tuning methods is provided in \cite{he2021towards}.}. 

\textbf{Model grafting}. By viewing the mismatch between numbers and text as a multimodal understanding problem, the notion of model grafting has emerged as another solution \cite{sun2021multilingual}. For example, in a recently proposed model called HeLM, non-text modalities are mapped via encoders (trained over input examples) into the same token embedding space as text, where a separate encoder is learned for each modality \cite{belyaeva2023multimodal}. In particular, HeLM trained two individual encoders to map spirogram sequences (lung capacity measurements) and demographic data into a limited-size token vocabulary that were inserted into the text tokens (being order-agnostic). Modality-specific encoders are also common in multimodal foundation models such as ImageBind \cite{girdhar2023imagebind} and Meta-Transformer \cite{zhang2023meta}, with the latter proposing a data-agnostic approach to merge token embeddings into a shared manifold space. In the same vein, ELIXR \cite{xu2023elixr}, LLaVA \cite{liu2023visual}, and BLIP2 \cite{li2023blip} introduced adapters\footnote{Term disambiguation: while adapters like LoRa insert trainable modules within layers, adapters like ELIXR correspond to independent models that plug into other pre-trained LLMs.} that mapped the activations of an image encoder into an LLM-understandable form. 

Model grafting has both advantages and disadvantages. On the one hand, it is computationally efficient when training the adapter layers, while allowing the LLM to connect to other high-performing models from different domains (e.g., a well-trained encoder of ECG data). Breaking down the problem into encoder, adapter, and LLM components can also facilitate faster testing and iteration cycles. On the other hand, this modularization introduces complexity by dealing with a system that is not trained end-to-end. Last, compared to natural language prompts, the communication between the specialist encoder (e.g., image or timeseries) and the LLM is no longer meaningful to humans since the mapping is represented as high-dimensional vectors \cite{googleblogMultimodalMedical}.

\textbf{New tokenizers}. Besides training lightweight adapters, we can rethink the design of tokenizers for mixed textual and numerical data. LLaMa was recently shown to outperform GPT-4 in arithmetic tasks because it splits each digit into an individual token, thereby ensuring consistent tokenization of numbers \cite{liu2023goat}. Specialized models trained on the scientific literature and notation also used similar digit-level splitting, pointing to improved understanding of symbolic and numerical data \cite{taylor2022galactica}.

\section{discussion}

This paper discussed the emerging idea of incorporating temporal data into pre-trained LLMs and proposed solutions for better encoding of numerical inputs into LLM-understandable information. We argue that this paradigm shift on how we leverage pre-trained models will particularly affect the area of ubiquitous computing, given the lack of modality-specific foundation models. To bridge the "modality gap", practitioners should focus on parameter-efficient transfer learning and model grafting, along with careful reassessment of existing tokenizers and embedders.  Beyond tokenization, the next generation of LLMs should address other architectural constraints such as longer context windows, towards better handling of real-world temporal datasets. Last, while our scope is limited to temporal data, these approaches could generalize to other non-language tasks \cite{dinh2022lift}.

\bibliographystyle{ACM-Reference-Format}
\bibliography{software}


\begin{thebibliography}{38}

%%% the Web address.

\ifx \showCODEN    \undefined \def \showCODEN     #1{\unskip}     \fi
\ifx \showDOI      \undefined \def \showDOI       #1{#1}\fi
\ifx \showISBNx    \undefined \def \showISBNx     #1{\unskip}     \fi
\ifx \showISBNxiii \undefined \def \showISBNxiii  #1{\unskip}     \fi
\ifx \showISSN     \undefined \def \showISSN      #1{\unskip}     \fi
\ifx \showLCCN     \undefined \def \showLCCN      #1{\unskip}     \fi
\ifx \shownote     \undefined \def \shownote      #1{#1}          \fi
\ifx \showarticletitle \undefined \def \showarticletitle #1{#1}   \fi
\ifx \showURL      \undefined \def \showURL       {\relax}        \fi
% invisible to TeX
\providecommand\bibfield[2]{#2}
\providecommand\bibinfo[2]{#2}
\providecommand\natexlab[1]{#1}
\providecommand\showeprint[2][]{arXiv:#2}

\bibitem[Belyaeva et~al\mbox{.}(2023)]%
        {belyaeva2023multimodal}
\bibfield{author}{\bibinfo{person}{Anastasiya Belyaeva},
  \bibinfo{person}{Justin Cosentino}, \bibinfo{person}{Farhad Hormozdiari},
  \bibinfo{person}{Cory~Y McLean}, {and} \bibinfo{person}{Nicholas~A
  Furlotte}.} \bibinfo{year}{2023}\natexlab{}.
\newblock \showarticletitle{Multimodal LLMs for health grounded in
  individual-specific data}.
\newblock \bibinfo{journal}{\emph{arXiv preprint arXiv:2307.09018}}
  (\bibinfo{year}{2023}).
\newblock


\bibitem[Bommasani et~al\mbox{.}(2021)]%
        {bommasani2021opportunities}
\bibfield{author}{\bibinfo{person}{Rishi Bommasani}, \bibinfo{person}{Drew~A
  Hudson}, \bibinfo{person}{Ehsan Adeli}, \bibinfo{person}{Russ Altman},
  \bibinfo{person}{Simran Arora}, \bibinfo{person}{Sydney von Arx},
  \bibinfo{person}{Michael~S Bernstein}, \bibinfo{person}{Jeannette Bohg},
  \bibinfo{person}{Antoine Bosselut}, \bibinfo{person}{Emma Brunskill},
  {et~al\mbox{.}}} \bibinfo{year}{2021}\natexlab{}.
\newblock \showarticletitle{On the opportunities and risks of foundation
  models}.
\newblock \bibinfo{journal}{\emph{arXiv preprint arXiv:2108.07258}}
  (\bibinfo{year}{2021}).
\newblock


\bibitem[Chowdhery et~al\mbox{.}(2022)]%
        {chowdhery2022palm}
\bibfield{author}{\bibinfo{person}{Aakanksha Chowdhery},
  \bibinfo{person}{Sharan Narang}, \bibinfo{person}{Jacob Devlin},
  \bibinfo{person}{Maarten Bosma}, \bibinfo{person}{Gaurav Mishra},
  \bibinfo{person}{Adam Roberts}, \bibinfo{person}{Paul Barham},
  \bibinfo{person}{Hyung~Won Chung}, \bibinfo{person}{Charles Sutton},
  \bibinfo{person}{Sebastian Gehrmann}, {et~al\mbox{.}}}
  \bibinfo{year}{2022}\natexlab{}.
\newblock \showarticletitle{Palm: Scaling language modeling with pathways}.
\newblock \bibinfo{journal}{\emph{arXiv preprint arXiv:2204.02311}}
  (\bibinfo{year}{2022}).
\newblock


\bibitem[Corrado and Matias({[n.\,d.]})]%
        {googleblogMultimodalMedical}
\bibfield{author}{\bibinfo{person}{Greg Corrado} {and} \bibinfo{person}{Yossi
  Matias}.} \bibinfo{year}{[n.\,d.]}\natexlab{}.
\newblock \bibinfo{title}{{M}ultimodal medical {A}{I} --- ai.googleblog.com}.
\newblock
  \bibinfo{howpublished}{\url{https://ai.googleblog.com/2023/08/multimodal-medical-ai.html}}.
\newblock
\newblock
\shownote{[Accessed 18-08-2023]}.


\bibitem[Dinh et~al\mbox{.}(2022)]%
        {dinh2022lift}
\bibfield{author}{\bibinfo{person}{Tuan Dinh}, \bibinfo{person}{Yuchen Zeng},
  \bibinfo{person}{Ruisu Zhang}, \bibinfo{person}{Ziqian Lin},
  \bibinfo{person}{Michael Gira}, \bibinfo{person}{Shashank Rajput},
  \bibinfo{person}{Jy-yong Sohn}, \bibinfo{person}{Dimitris Papailiopoulos},
  {and} \bibinfo{person}{Kangwook Lee}.} \bibinfo{year}{2022}\natexlab{}.
\newblock \showarticletitle{Lift: Language-interfaced fine-tuning for
  non-language machine learning tasks}.
\newblock \bibinfo{journal}{\emph{Advances in Neural Information Processing
  Systems}}  \bibinfo{volume}{35} (\bibinfo{year}{2022}),
  \bibinfo{pages}{11763--11784}.
\newblock


\bibitem[Gaudilliere et~al\mbox{.}(2021)]%
        {gaudilliere2021generative}
\bibfield{author}{\bibinfo{person}{Pierre~Louis Gaudilliere},
  \bibinfo{person}{Halla Sigurthorsdottir}, \bibinfo{person}{Cl{\'e}mentine
  Aguet}, \bibinfo{person}{J{\'e}r{\^o}me Van~Zaen}, \bibinfo{person}{Mathieu
  Lemay}, {and} \bibinfo{person}{Ricard Delgado-Gonzalo}.}
  \bibinfo{year}{2021}\natexlab{}.
\newblock \showarticletitle{Generative Pre-Trained Transformer for Cardiac
  Abnormality Detection}. In \bibinfo{booktitle}{\emph{2021 Computing in
  Cardiology (CinC)}}, Vol.~\bibinfo{volume}{48}. IEEE, \bibinfo{pages}{1--4}.
\newblock


\bibitem[Girdhar et~al\mbox{.}(2023)]%
        {girdhar2023imagebind}
\bibfield{author}{\bibinfo{person}{Rohit Girdhar}, \bibinfo{person}{Alaaeldin
  El-Nouby}, \bibinfo{person}{Zhuang Liu}, \bibinfo{person}{Mannat Singh},
  \bibinfo{person}{Kalyan~Vasudev Alwala}, \bibinfo{person}{Armand Joulin},
  {and} \bibinfo{person}{Ishan Misra}.} \bibinfo{year}{2023}\natexlab{}.
\newblock \showarticletitle{Imagebind: One embedding space to bind them all}.
  In \bibinfo{booktitle}{\emph{Proceedings of the IEEE/CVF Conference on
  Computer Vision and Pattern Recognition}}. \bibinfo{pages}{15180--15190}.
\newblock


\bibitem[Haresamudram et~al\mbox{.}(2022)]%
        {haresamudram2022assessing}
\bibfield{author}{\bibinfo{person}{Harish Haresamudram}, \bibinfo{person}{Irfan
  Essa}, {and} \bibinfo{person}{Thomas Pl{\"o}tz}.}
  \bibinfo{year}{2022}\natexlab{}.
\newblock \showarticletitle{Assessing the state of self-supervised human
  activity recognition using wearables}.
\newblock \bibinfo{journal}{\emph{Proceedings of the ACM on Interactive,
  Mobile, Wearable and Ubiquitous Technologies}} \bibinfo{volume}{6},
  \bibinfo{number}{3} (\bibinfo{year}{2022}), \bibinfo{pages}{1--47}.
\newblock


\bibitem[He et~al\mbox{.}(2021)]%
        {he2021towards}
\bibfield{author}{\bibinfo{person}{Junxian He}, \bibinfo{person}{Chunting
  Zhou}, \bibinfo{person}{Xuezhe Ma}, \bibinfo{person}{Taylor
  Berg-Kirkpatrick}, {and} \bibinfo{person}{Graham Neubig}.}
  \bibinfo{year}{2021}\natexlab{}.
\newblock \showarticletitle{Towards a unified view of parameter-efficient
  transfer learning}.
\newblock \bibinfo{journal}{\emph{arXiv preprint arXiv:2110.04366}}
  (\bibinfo{year}{2021}).
\newblock


\bibitem[Hu et~al\mbox{.}(2021)]%
        {hu2021lora}
\bibfield{author}{\bibinfo{person}{Edward~J Hu}, \bibinfo{person}{Yelong Shen},
  \bibinfo{person}{Phillip Wallis}, \bibinfo{person}{Zeyuan Allen-Zhu},
  \bibinfo{person}{Yuanzhi Li}, \bibinfo{person}{Shean Wang},
  \bibinfo{person}{Lu Wang}, {and} \bibinfo{person}{Weizhu Chen}.}
  \bibinfo{year}{2021}\natexlab{}.
\newblock \showarticletitle{Lora: Low-rank adaptation of large language
  models}.
\newblock \bibinfo{journal}{\emph{arXiv preprint arXiv:2106.09685}}
  (\bibinfo{year}{2021}).
\newblock


\bibitem[Kudo and Richardson(2018)]%
        {kudo2018sentencepiece}
\bibfield{author}{\bibinfo{person}{Taku Kudo} {and} \bibinfo{person}{John
  Richardson}.} \bibinfo{year}{2018}\natexlab{}.
\newblock \showarticletitle{Sentencepiece: A simple and language independent
  subword tokenizer and detokenizer for neural text processing}.
\newblock \bibinfo{journal}{\emph{arXiv preprint arXiv:1808.06226}}
  (\bibinfo{year}{2018}).
\newblock


\bibitem[Lester et~al\mbox{.}(2021)]%
        {lester2021power}
\bibfield{author}{\bibinfo{person}{Brian Lester}, \bibinfo{person}{Rami
  Al-Rfou}, {and} \bibinfo{person}{Noah Constant}.}
  \bibinfo{year}{2021}\natexlab{}.
\newblock \showarticletitle{The power of scale for parameter-efficient prompt
  tuning}.
\newblock \bibinfo{journal}{\emph{arXiv preprint arXiv:2104.08691}}
  (\bibinfo{year}{2021}).
\newblock


\bibitem[Li et~al\mbox{.}(2023)]%
        {li2023blip}
\bibfield{author}{\bibinfo{person}{Junnan Li}, \bibinfo{person}{Dongxu Li},
  \bibinfo{person}{Silvio Savarese}, {and} \bibinfo{person}{Steven Hoi}.}
  \bibinfo{year}{2023}\natexlab{}.
\newblock \showarticletitle{Blip-2: Bootstrapping language-image pre-training
  with frozen image encoders and large language models}.
\newblock \bibinfo{journal}{\emph{arXiv preprint arXiv:2301.12597}}
  (\bibinfo{year}{2023}).
\newblock


\bibitem[Liang et~al\mbox{.}(2022)]%
        {liang2022mind}
\bibfield{author}{\bibinfo{person}{Victor~Weixin Liang}, \bibinfo{person}{Yuhui
  Zhang}, \bibinfo{person}{Yongchan Kwon}, \bibinfo{person}{Serena Yeung},
  {and} \bibinfo{person}{James~Y Zou}.} \bibinfo{year}{2022}\natexlab{}.
\newblock \showarticletitle{Mind the gap: Understanding the modality gap in
  multi-modal contrastive representation learning}.
\newblock \bibinfo{journal}{\emph{Advances in Neural Information Processing
  Systems}}  \bibinfo{volume}{35} (\bibinfo{year}{2022}),
  \bibinfo{pages}{17612--17625}.
\newblock


\bibitem[Liu et~al\mbox{.}(2023a)]%
        {liu2023visual}
\bibfield{author}{\bibinfo{person}{Haotian Liu}, \bibinfo{person}{Chunyuan Li},
  \bibinfo{person}{Qingyang Wu}, {and} \bibinfo{person}{Yong~Jae Lee}.}
  \bibinfo{year}{2023}\natexlab{a}.
\newblock \showarticletitle{Visual instruction tuning}.
\newblock \bibinfo{journal}{\emph{arXiv preprint arXiv:2304.08485}}
  (\bibinfo{year}{2023}).
\newblock


\bibitem[Liu and Low(2023)]%
        {liu2023goat}
\bibfield{author}{\bibinfo{person}{Tiedong Liu} {and} \bibinfo{person}{Bryan
  Kian~Hsiang Low}.} \bibinfo{year}{2023}\natexlab{}.
\newblock \bibinfo{title}{Goat: Fine-tuned LLaMA Outperforms GPT-4 on
  Arithmetic Tasks}.
\newblock
\newblock
\showeprint[arxiv]{2305.14201}~[cs.LG]


\bibitem[Liu et~al\mbox{.}(2023b)]%
        {liu2023large}
\bibfield{author}{\bibinfo{person}{Xin Liu}, \bibinfo{person}{Daniel McDuff},
  \bibinfo{person}{Geza Kovacs}, \bibinfo{person}{Isaac Galatzer-Levy},
  \bibinfo{person}{Jacob Sunshine}, \bibinfo{person}{Jiening Zhan},
  \bibinfo{person}{Ming-Zher Poh}, \bibinfo{person}{Shun Liao},
  \bibinfo{person}{Paolo Di~Achille}, {and} \bibinfo{person}{Shwetak Patel}.}
  \bibinfo{year}{2023}\natexlab{b}.
\newblock \showarticletitle{Large Language Models are Few-Shot Health
  Learners}.
\newblock \bibinfo{journal}{\emph{arXiv preprint arXiv:2305.15525}}
  (\bibinfo{year}{2023}).
\newblock


\bibitem[Lu et~al\mbox{.}(2021)]%
        {lu2021pretrained}
\bibfield{author}{\bibinfo{person}{Kevin Lu}, \bibinfo{person}{Aditya Grover},
  \bibinfo{person}{Pieter Abbeel}, {and} \bibinfo{person}{Igor Mordatch}.}
  \bibinfo{year}{2021}\natexlab{}.
\newblock \showarticletitle{Pretrained transformers as universal computation
  engines}.
\newblock \bibinfo{journal}{\emph{arXiv preprint arXiv:2103.05247}}
  \bibinfo{volume}{1} (\bibinfo{year}{2021}).
\newblock


\bibitem[Ma et~al\mbox{.}(2019)]%
        {ma2019attnsense}
\bibfield{author}{\bibinfo{person}{Haojie Ma}, \bibinfo{person}{Wenzhong Li},
  \bibinfo{person}{Xiao Zhang}, \bibinfo{person}{Songcheng Gao}, {and}
  \bibinfo{person}{Sanglu Lu}.} \bibinfo{year}{2019}\natexlab{}.
\newblock \showarticletitle{AttnSense: Multi-level attention mechanism for
  multimodal human activity recognition.}. In
  \bibinfo{booktitle}{\emph{IJCAI}}. \bibinfo{pages}{3109--3115}.
\newblock


\bibitem[Millidge({[n.\,d.]})]%
        {berenIntegerTokenization}
\bibfield{author}{\bibinfo{person}{Beren Millidge}.}
  \bibinfo{year}{[n.\,d.]}\natexlab{}.
\newblock \bibinfo{title}{{I}nteger tokenization is insane --- beren.io}.
\newblock
  \bibinfo{howpublished}{\url{https://www.beren.io/2023-02-04-Integer-tokenization-is-insane/.}}.
\newblock
\newblock
\shownote{[Accessed 18-08-2023]}.


\bibitem[Nogueira et~al\mbox{.}(2021)]%
        {nogueira2021investigating}
\bibfield{author}{\bibinfo{person}{Rodrigo Nogueira}, \bibinfo{person}{Zhiying
  Jiang}, {and} \bibinfo{person}{Jimmy Lin}.} \bibinfo{year}{2021}\natexlab{}.
\newblock \showarticletitle{Investigating the limitations of transformers with
  simple arithmetic tasks}.
\newblock \bibinfo{journal}{\emph{arXiv preprint arXiv:2102.13019}}
  (\bibinfo{year}{2021}).
\newblock


\bibitem[OpenAI(2023)]%
        {openai2023gpt}
\bibfield{author}{\bibinfo{person}{OpenAI}.} \bibinfo{year}{2023}\natexlab{}.
\newblock \showarticletitle{GPT-4 technical report}.
\newblock \bibinfo{journal}{\emph{arXiv}} (\bibinfo{year}{2023}),
  \bibinfo{pages}{2303--08774}.
\newblock


\bibitem[Saeed et~al\mbox{.}(2019)]%
        {saeed2019multi}
\bibfield{author}{\bibinfo{person}{Aaqib Saeed}, \bibinfo{person}{Tanir
  Ozcelebi}, {and} \bibinfo{person}{Johan Lukkien}.}
  \bibinfo{year}{2019}\natexlab{}.
\newblock \showarticletitle{Multi-task self-supervised learning for human
  activity detection}.
\newblock \bibinfo{journal}{\emph{Proceedings of the ACM on Interactive,
  Mobile, Wearable and Ubiquitous Technologies}} \bibinfo{volume}{3},
  \bibinfo{number}{2} (\bibinfo{year}{2019}), \bibinfo{pages}{1--30}.
\newblock


\bibitem[Schuster and Nakajima(2012)]%
        {schuster2012japanese}
\bibfield{author}{\bibinfo{person}{Mike Schuster} {and}
  \bibinfo{person}{Kaisuke Nakajima}.} \bibinfo{year}{2012}\natexlab{}.
\newblock \showarticletitle{Japanese and korean voice search}. In
  \bibinfo{booktitle}{\emph{2012 IEEE international conference on acoustics,
  speech and signal processing (ICASSP)}}. IEEE, \bibinfo{pages}{5149--5152}.
\newblock


\bibitem[Sennrich et~al\mbox{.}(2016)]%
        {sennrich2016neural}
\bibfield{author}{\bibinfo{person}{Rico Sennrich}, \bibinfo{person}{Barry
  Haddow}, {and} \bibinfo{person}{Alexandra Birch}.}
  \bibinfo{year}{2016}\natexlab{}.
\newblock \showarticletitle{Neural Machine Translation of Rare Words with
  Subword Units}. In \bibinfo{booktitle}{\emph{54th Annual Meeting of the
  Association for Computational Linguistics}}. Association for Computational
  Linguistics (ACL), \bibinfo{pages}{1715--1725}.
\newblock


\bibitem[Shi et~al\mbox{.}(2023)]%
        {shi2023towards}
\bibfield{author}{\bibinfo{person}{Peiyang Shi}, \bibinfo{person}{Michael~C
  Welle}, \bibinfo{person}{M{\aa}rten Bj{\"o}rkman}, {and}
  \bibinfo{person}{Danica Kragic}.} \bibinfo{year}{2023}\natexlab{}.
\newblock \showarticletitle{Towards understanding the modality gap in CLIP}. In
  \bibinfo{booktitle}{\emph{ICLR 2023 Workshop on Multimodal Representation
  Learning: Perks and Pitfalls}}.
\newblock


\bibitem[Sooriya~Patabandige et~al\mbox{.}(2023)]%
        {sooriya2023poster}
\bibfield{author}{\bibinfo{person}{Pramuka~Medaranga Sooriya~Patabandige},
  \bibinfo{person}{Steven Antya~Orvala Waskito}, \bibinfo{person}{Kunjun Li},
  \bibinfo{person}{Kai~Jie Leow}, \bibinfo{person}{Shantanu Chakrabarty}, {and}
  \bibinfo{person}{Ambuj Varshney}.} \bibinfo{year}{2023}\natexlab{}.
\newblock \showarticletitle{Poster: Rethinking Embedded Sensor Data Processing
  and Analysis with Large Language Models}. In
  \bibinfo{booktitle}{\emph{Proceedings of the 21st Annual International
  Conference on Mobile Systems, Applications and Services}}.
  \bibinfo{pages}{561--562}.
\newblock


\bibitem[Spathis et~al\mbox{.}(2021)]%
        {spathis2021self}
\bibfield{author}{\bibinfo{person}{Dimitris Spathis}, \bibinfo{person}{Ignacio
  Perez-Pozuelo}, \bibinfo{person}{Soren Brage}, \bibinfo{person}{Nicholas~J
  Wareham}, {and} \bibinfo{person}{Cecilia Mascolo}.}
  \bibinfo{year}{2021}\natexlab{}.
\newblock \showarticletitle{Self-supervised transfer learning of physiological
  representations from free-living wearable data}. In
  \bibinfo{booktitle}{\emph{Proceedings of the Conference on Health, Inference,
  and Learning}}. \bibinfo{pages}{69--78}.
\newblock


\bibitem[Spathis et~al\mbox{.}(2022)]%
        {spathis2022breaking}
\bibfield{author}{\bibinfo{person}{Dimitris Spathis}, \bibinfo{person}{Ignacio
  Perez-Pozuelo}, \bibinfo{person}{Laia Marques-Fernandez}, {and}
  \bibinfo{person}{Cecilia Mascolo}.} \bibinfo{year}{2022}\natexlab{}.
\newblock \showarticletitle{Breaking away from labels: The promise of
  self-supervised machine learning in intelligent health}.
\newblock \bibinfo{journal}{\emph{Patterns}} \bibinfo{volume}{3},
  \bibinfo{number}{2} (\bibinfo{year}{2022}).
\newblock


\bibitem[Sun et~al\mbox{.}(2021)]%
        {sun2021multilingual}
\bibfield{author}{\bibinfo{person}{Zewei Sun}, \bibinfo{person}{Mingxuan Wang},
  {and} \bibinfo{person}{Lei Li}.} \bibinfo{year}{2021}\natexlab{}.
\newblock \showarticletitle{Multilingual translation via grafting pre-trained
  language models}.
\newblock \bibinfo{journal}{\emph{arXiv preprint arXiv:2109.05256}}
  (\bibinfo{year}{2021}).
\newblock


\bibitem[Tang et~al\mbox{.}(2021)]%
        {tang2021selfhar}
\bibfield{author}{\bibinfo{person}{Chi~Ian Tang}, \bibinfo{person}{Ignacio
  Perez-Pozuelo}, \bibinfo{person}{Dimitris Spathis}, \bibinfo{person}{Soren
  Brage}, \bibinfo{person}{Nick Wareham}, {and} \bibinfo{person}{Cecilia
  Mascolo}.} \bibinfo{year}{2021}\natexlab{}.
\newblock \showarticletitle{Selfhar: Improving human activity recognition
  through self-training with unlabeled data}.
\newblock \bibinfo{journal}{\emph{Proceedings of the ACM on interactive,
  mobile, wearable and ubiquitous technologies}} \bibinfo{volume}{5},
  \bibinfo{number}{1} (\bibinfo{year}{2021}), \bibinfo{pages}{1--30}.
\newblock


\bibitem[Taylor et~al\mbox{.}(2022)]%
        {taylor2022galactica}
\bibfield{author}{\bibinfo{person}{Ross Taylor}, \bibinfo{person}{Marcin
  Kardas}, \bibinfo{person}{Guillem Cucurull}, \bibinfo{person}{Thomas
  Scialom}, \bibinfo{person}{Anthony Hartshorn}, \bibinfo{person}{Elvis
  Saravia}, \bibinfo{person}{Andrew Poulton}, \bibinfo{person}{Viktor Kerkez},
  {and} \bibinfo{person}{Robert Stojnic}.} \bibinfo{year}{2022}\natexlab{}.
\newblock \bibinfo{title}{Galactica: A Large Language Model for Science}.
\newblock
\newblock
\showeprint[arxiv]{2211.09085}~[cs.CL]


\bibitem[Vaid et~al\mbox{.}(2023)]%
        {vaid2023foundational}
\bibfield{author}{\bibinfo{person}{Akhil Vaid}, \bibinfo{person}{Joy Jiang},
  \bibinfo{person}{Ashwin Sawant}, \bibinfo{person}{Stamatios Lerakis},
  \bibinfo{person}{Edgar Argulian}, \bibinfo{person}{Yuri Ahuja},
  \bibinfo{person}{Joshua Lampert}, \bibinfo{person}{Alexander Charney},
  \bibinfo{person}{Hayit Greenspan}, \bibinfo{person}{Jagat Narula},
  {et~al\mbox{.}}} \bibinfo{year}{2023}\natexlab{}.
\newblock \showarticletitle{A foundational vision transformer improves
  diagnostic performance for electrocardiograms}.
\newblock \bibinfo{journal}{\emph{NPJ Digital Medicine}} \bibinfo{volume}{6},
  \bibinfo{number}{1} (\bibinfo{year}{2023}), \bibinfo{pages}{108}.
\newblock


\bibitem[Weiss et~al\mbox{.}(2019)]%
        {weiss2019smartphone}
\bibfield{author}{\bibinfo{person}{Gary~M Weiss}, \bibinfo{person}{Kenichi
  Yoneda}, {and} \bibinfo{person}{Thaier Hayajneh}.}
  \bibinfo{year}{2019}\natexlab{}.
\newblock \showarticletitle{Smartphone and smartwatch-based biometrics using
  activities of daily living}.
\newblock \bibinfo{journal}{\emph{IEEE Access}}  \bibinfo{volume}{7}
  (\bibinfo{year}{2019}), \bibinfo{pages}{133190--133202}.
\newblock


\bibitem[Xu et~al\mbox{.}(2023)]%
        {xu2023elixr}
\bibfield{author}{\bibinfo{person}{Shawn Xu}, \bibinfo{person}{Lin Yang},
  \bibinfo{person}{Christopher Kelly}, \bibinfo{person}{Marcin Sieniek},
  \bibinfo{person}{Timo Kohlberger}, \bibinfo{person}{Martin Ma},
  \bibinfo{person}{Wei-Hung Weng}, \bibinfo{person}{Attila Kiraly},
  \bibinfo{person}{Sahar Kazemzadeh}, \bibinfo{person}{Zakkai Melamed},
  \bibinfo{person}{Jungyeon Park}, \bibinfo{person}{Patricia Strachan},
  \bibinfo{person}{Yun Liu}, \bibinfo{person}{Chuck Lau},
  \bibinfo{person}{Preeti Singh}, \bibinfo{person}{Christina Chen},
  \bibinfo{person}{Mozziyar Etemadi}, \bibinfo{person}{Sreenivasa~Raju
  Kalidindi}, \bibinfo{person}{Yossi Matias}, \bibinfo{person}{Katherine Chou},
  \bibinfo{person}{Greg~S. Corrado}, \bibinfo{person}{Shravya Shetty},
  \bibinfo{person}{Daniel Tse}, \bibinfo{person}{Shruthi Prabhakara},
  \bibinfo{person}{Daniel Golden}, \bibinfo{person}{Rory Pilgrim},
  \bibinfo{person}{Krish Eswaran}, {and} \bibinfo{person}{Andrew Sellergren}.}
  \bibinfo{year}{2023}\natexlab{}.
\newblock \bibinfo{title}{ELIXR: Towards a general purpose X-ray artificial
  intelligence system through alignment of large language models and radiology
  vision encoders}.
\newblock
\newblock
\showeprint[arxiv]{2308.01317}~[cs.CV]


\bibitem[Xue and Salim(2022)]%
        {xue2022prompt}
\bibfield{author}{\bibinfo{person}{Hao Xue} {and} \bibinfo{person}{Flora~D
  Salim}.} \bibinfo{year}{2022}\natexlab{}.
\newblock \showarticletitle{Prompt-Based Time Series Forecasting: A New Task
  and Dataset}.
\newblock \bibinfo{journal}{\emph{arXiv preprint arXiv:2210.08964}}
  (\bibinfo{year}{2022}).
\newblock


\bibitem[Yuan et~al\mbox{.}(2022)]%
        {yuan2022self}
\bibfield{author}{\bibinfo{person}{Hang Yuan}, \bibinfo{person}{Shing Chan},
  \bibinfo{person}{Andrew~P Creagh}, \bibinfo{person}{Catherine Tong},
  \bibinfo{person}{David~A Clifton}, {and} \bibinfo{person}{Aiden Doherty}.}
  \bibinfo{year}{2022}\natexlab{}.
\newblock \showarticletitle{Self-supervised learning for human activity
  recognition using 700,000 person-days of wearable data}.
\newblock \bibinfo{journal}{\emph{arXiv preprint arXiv:2206.02909}}
  (\bibinfo{year}{2022}).
\newblock


\bibitem[Zhang et~al\mbox{.}(2023)]%
        {zhang2023meta}
\bibfield{author}{\bibinfo{person}{Yiyuan Zhang}, \bibinfo{person}{Kaixiong
  Gong}, \bibinfo{person}{Kaipeng Zhang}, \bibinfo{person}{Hongsheng Li},
  \bibinfo{person}{Yu Qiao}, \bibinfo{person}{Wanli Ouyang}, {and}
  \bibinfo{person}{Xiangyu Yue}.} \bibinfo{year}{2023}\natexlab{}.
\newblock \showarticletitle{Meta-Transformer: A Unified Framework for
  Multimodal Learning}.
\newblock \bibinfo{journal}{\emph{arXiv preprint arXiv:2307.10802}}
  (\bibinfo{year}{2023}).
\newblock


\end{thebibliography}

\end{document}